\documentclass[sigconf]{acmart}
\AtBeginDocument{%
  }

\setcopyright{acmlicensed}
\copyrightyear{2025}
\acmYear{2018}
\acmDOI{XXXXXXX.XXXXXXX}
\acmConference[Conference acronym 'XX]{Make sure to enter the correct
  conference title from your rights confirmation email}{June 03--05,
  2018}{Woodstock, NY}
\acmISBN{978-1-4503-XXXX-X/2018/06}

\usepackage{graphicx}
\usepackage{subcaption}
\usepackage{caption}
\begin{document}

\title{Controllable Generation of Large-Scale 3D Urban Layouts with Semantic and Structural Guidance}

\author{Mengyuan Niu}
\orcid{0009-0009-9269-9560}
\affiliation{%
  \institution{Southeast University}
  \city{Nanjing}
  \state{Jiangsu}
  \country{China}
}
\email{myniu@seu.edu.cn}

\author{Xinxin Zhuo}
\affiliation{%
  \institution{Southeast University}
  \city{Nanjing}
  \state{Jiangsu}
  \country{China}}
\email{zhuoxinxin@seu.edu.cn}

\author{Ruizhe Wang}
\affiliation{%
  \institution{Southeast University}
  \city{Nanjing}
  \state{Jiangsu}
  \country{China}}
\email{rz_wang@seu.edu.cn}

\author{Yuyue Huang}
\affiliation{%
  \institution{Southeast University}
  \city{Nanjing}
  \state{Jiangsu}
  \country{China}}
\email{huangyy0628@163.com}

\author{Junyan Yang}
\affiliation{%
  \institution{Southeast University}
  \city{Nanjing}
  \state{Jiangsu}
  \country{China}}
\email{yangjy_seu@163.com}

\author{Qiao Wang}
\affiliation{%
  \institution{Southeast University}
  \city{Nanjing}
  \state{Jiangsu}
  \country{China}}
\email{qiaowang@seu.edu.cn}

% \affiliation{%

% \affiliation{%

\renewcommand{\shortauthors}{Trovato et al.}

\begin{abstract}

%Urban modeling and generation play a crucial role in city planning, scene synthesis, and game industry. Previous image-based methods can generate diverse layout variations but often fail to preserve geometric continuity and are unsuitable for producing large-scale urban vector data. Although graph-based approaches naturally capture structural relationships, they typically overlook the semantic attributes of individual blocks. In this paper we introduce a controllable generative framework for large-scale 3D vector urban layout design conditioned on given geometry and semantic attributes. Our work fuse geometric and semantic parcel attributes to guide building generation, thereby enhancing diversity and controllability. We also introduce edge weights to better capture layout patterns. Additionally, by embedding building height into the graph construction, we extend generation and design from 2D to realistic and plausible 3D urban structures. Experiments demonstrate that our method produces valid, large-scale 3D vector base models of urban layouts, offering a powerful tool for data-driven city planning and design.

Urban modeling is essential for city planning, scene synthesis, and gaming. Existing image-based methods generate diverse layouts but often lack geometric continuity and scalability, while graph-based methods capture structural relations yet overlook parcel semantics. We present a controllable framework for large-scale 3D vector urban layout generation, conditioned on both geometry and semantics. By fusing geometric and semantic attributes, introducing edge weights, and embedding building height in the graph, our method extends 2D layouts to realistic 3D structures. It also enables users to directly control the output by modifying semantic attributes. Experiments show that it produces valid, large-scale urban models, offering an effective tool for data-driven planning and design.

  % A clear and well-documented \LaTeX\ document is presented as an
  % article formatted for publication by ACM in a conference proceedings
  % or journal publication. Based on the ``acmart'' document class, this
  % article presents and explains many of the common variations, as well
  % as many of the formatting elements an author may use in the
  % preparation of the documentation of their work.
\end{abstract}


\begin{CCSXML}
<ccs2012>
   <concept>
       <concept_id>10010405.10010469.10010472.10010440</concept_id>
       <concept_desc>Applied computing~Computer-aided design</concept_desc>
       <concept_significance>500</concept_significance>
       </concept>
   <concept>
       <concept_id>10010147.10010178.10010224.10010225.10010227</concept_id>
       <concept_desc>Computing methodologies~Scene understanding</concept_desc>
       <concept_significance>500</concept_significance>
       </concept>
 </ccs2012>
\end{CCSXML}

\ccsdesc[500]{Applied computing~Computer-aided design}
\ccsdesc[500]{Computing methodologies~Scene understanding}

% \keywords{Do, Not, Use, This, Code, Put, the, Correct, Terms, for,
\keywords{City Planning, Layout Generation, Building Representation, Graph Network}

\begin{teaserfigure}
  \centering
  %—— 第二行第五张子图 ——  
  \begin{subfigure}[b]{0.8\textwidth}
    \includegraphics[width=\textwidth]{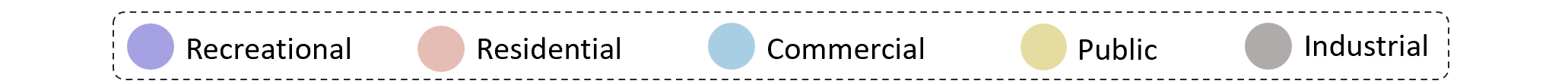}
    % \caption{Your extra caption}
    \label{fig:teaser5}
  \end{subfigure}
  %—— 第一行四张子图 ——  
  \begin{subfigure}[b]{0.24\textwidth}
    \includegraphics[width=\textwidth]{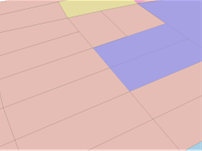}
    \caption{Given block boundary and landuse}
    \label{fig:teaser1}
  \end{subfigure}\hfill
  \begin{subfigure}[b]{0.24\textwidth}
    \includegraphics[width=\textwidth]{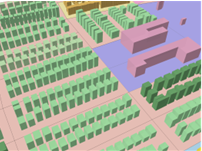}
    \caption{Generated fundamental 3D Layout}
    \label{fig:teaser2}
  \end{subfigure}\hfill
  \begin{subfigure}[b]{0.24\textwidth}
    \includegraphics[width=\textwidth]{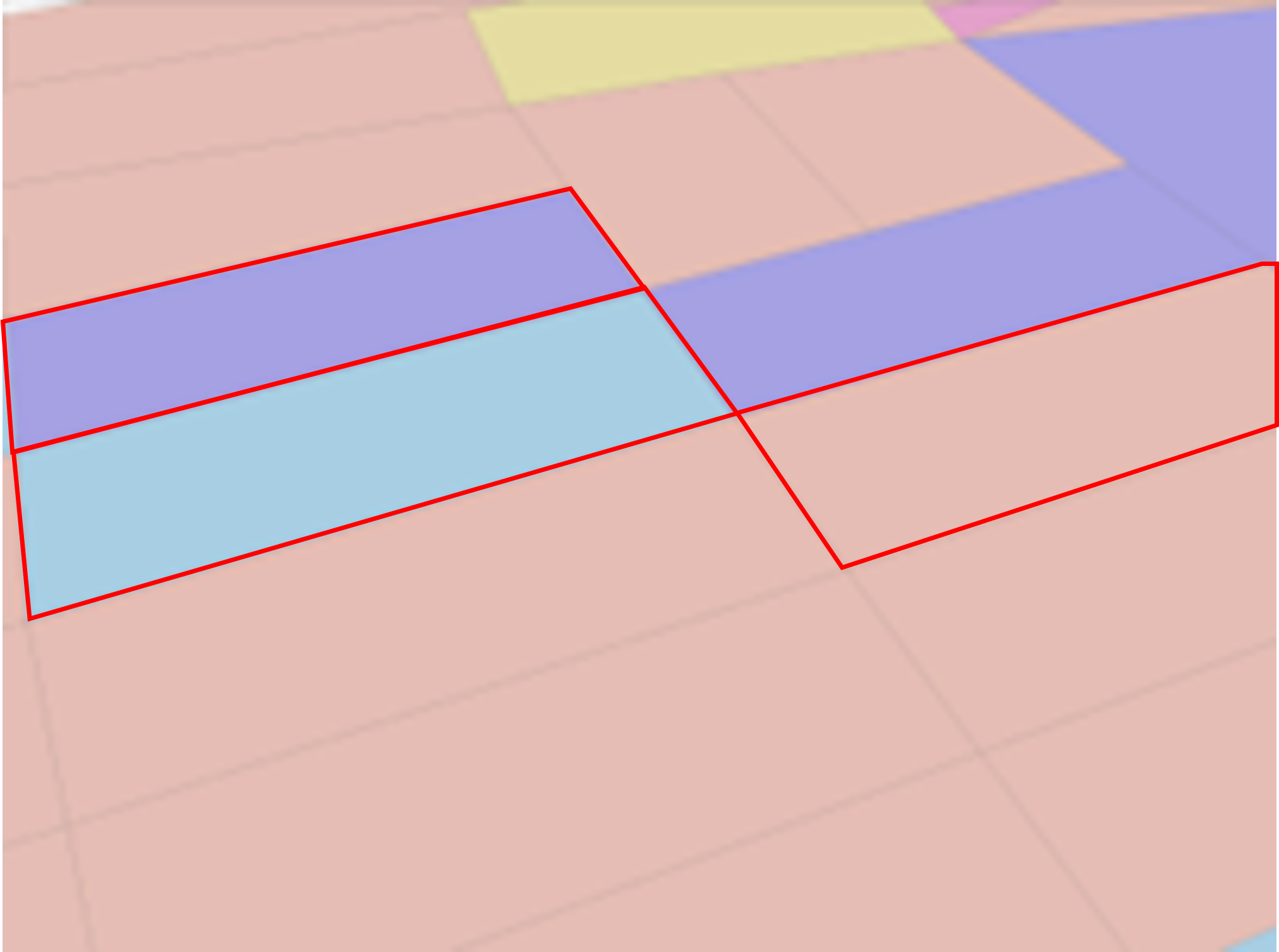}
    \caption{Change landuse type of several blocks}
    \label{fig:teaser3}
  \end{subfigure}\hfill
  \begin{subfigure}[b]{0.24\textwidth}
    \includegraphics[width=\textwidth]{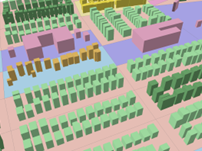}
    \caption{New 3D layout within specific blocks}
    \label{fig:teaser4}
  \end{subfigure} 
  \caption{Foundational 3D city generation with geometry and semantic information. Our method generates realistic urban layouts given the block boundary and landuse type (Left Pair). Moreover, new functional urban layouts will be generated only with change of the landuse type within different regions (Right Pair). The colors of the blocks indicate distinct land-use categories, detailed in the upper diagram.}
  \Description{there are four pics illustrating different content}
  \label{fig:urban_layout}
\end{teaserfigure}

\maketitle

\section{Introduction}
City modeling has attracted sustained interest across domains such as computer graphics, urban planning, and game development \cite{ebert2003procedural, cureton2023gamebased, ghisleni2023urbanblocks}. Cities are increasingly conceptualized as integrated systems comprising roads, buildings, water bodies, and vegetation. Among these, road networks delineate land parcels that form urban blocks—fundamental spatial units that structure urban form and function \cite{shakibamanesh2020urbanblocks, gasparovic2021transformations}.

Traditional procedural modeling approaches generate large-scale urban environments using rule-based grammars and templates \cite{benes2011guided, watson2008procedural}. While effective for regular grid-aligned layouts, they struggle to accommodate irregular geometries and heterogeneous architectural styles. Recent data-driven methods address these limitations by leveraging deep generative models. Pix2Pix, for instance, conditions on road and parcel inputs to synthesize building distributions, though often with blurred boundaries \cite{isola2017pix2pix, li2024mapping}. Extensions such as ESGAN and CAIN-GAN improve output diversity and controllability \cite{esgan2023github, jiang2024cain}, while diffusion-based frameworks like Stable Diffusion and ControlNet enable layout generation from semantic maps or bird’s-eye-view (BEV) inputs \cite{zhang2023controlnet, skydiffusion2024}. GeoDiffusion further encodes geographic primitives as textual prompts to synthesize high-fidelity BEV layouts \cite{geodiffusion2024}.

Parallel to image-based approaches, vector-graph-based methods represent urban environments as graph-structured data\cite{johnson2024scalable} \cite{lee2025vector}, enabling fine-grained control and semantic reasoning. LayoutVAE learns distributions over bounding boxes of buildings and roads via variational autoencoding \cite{jyothi2019layoutvae}. BlockPlanner encodes urban blocks as graphs with parcel-level nodes and adjacency edges, using graph neural networks to generate regular footprints \cite{xu2021blockplanner}. GlobalMapper generalizes this to arbitrary polygonal parcels using graph attention networks, supporting conditional control over semantic and geometric attributes \cite{he2023globalmapper}. Complementary efforts have explored semantic-aware synthesis \cite{kelly2018cityengine}, multi-modal integration of geometry and land use \cite{biljecki2018overview}, and high-fidelity façade generation at the building level \cite{li2021citydreamer}. These developments underscore the need for scalable frameworks that unify geometric realism, semantic consistency, and topological coherence in urban layout generation.

However, existing building layout optimization focuses on 2D planes and arranges parcels solely by shape, overlooking real-world attributes such as parcel scale, orientation, and semantic context. Image-based 3D city generation methods can produce high-fidelity local scenes but struggle to scale to large-area planning, maintain spatial continuity, and avoid pixel‐induced jagged edges that contradict authentic urban design. Semantic maps and depth maps are often generated separately, failing to ensure consistent height and semantic alignment, and the outputs lack vector structure, limiting controllability and downstream editing.

To address these challenges, we propose a unified framework which can generate diverse, controllable, and realistic urban layouts as foundational vector data. Our approach treats each building as an independent entity rather than a collection of pixels, inherently producing 3D vectorized outputs. By jointly modeling geometric and semantic factors in the generation process, we achieve more plausible and realistic layouts. Moreover, our framework can synthesize layouts at arbitrary scales from a single input—unbounded by image resolution—thereby preserving spatial continuity.

To effectively represent and model urban blocks and their buildings, we first employ a unified graph representation of block–building relationships\cite{he2023globalmapper}. It's noteworthy that we introduce an adaptive edge weights to enrich this structure and allow more efficient propagation of node features, based on our observation that buildings tend to align in grid-like rows parallel to parcel boundaries. Next, we abstract individual buildings as basic 3D solids defined by their footprint geometries and height profiles. Treating each building as an integrated unit ensures plausibility in the generated results. By adopting a Graph Attention Network (GAT)\cite{velickovic2018gat} as the backbone of our Conditional Variational Autoencoder (CVAE)\cite{sohn2015cvae}, the model learns height and layout patterns simultaneously, achieving 3D planning results that closely approximate real-world cities. Finally, we fuse parcel semantic labels and geometric attributes into a single conditional signal that supervises both the learning and the generation of urban layouts, ensuring that the outputs are reasonable and effective. 

In addition, our method supports user-driven edits of land use attributes to produce various planning proposals, and the resulting base vector models support immediate editing, rendering, detailed refinement, and seamless alignment with existing GIS datasets, greatly enhancing both usability and practical application.Our main contributions can be summarized as follows:
\begin{itemize}
\item We propose a novel framework for generating city-scale building layouts conditioned on parcel geometric and semantic attributes.
\item We propose a strengthed graph representation that integrates row–column layout patterns into edge weights and fuses multiple parcel attributes as conditional inputs, yielding greater diversity and controllability.
\item We embed building height as a node feature to generate continuous, realistic, vector-based 3D urban building models at arbitrary scales.

\end{itemize}

\section{Related Work}

\subsection{Urban Layout Generation}
Early research in urban layout generation predominantly relied on procedural modeling frameworks, leveraging shape grammars and production rules to encode hierarchical street networks and block structures~\cite{parish2001procedural,muller2006procedural,weber2009interactive}. These methods offer fine-grained control over stylistic variation but require labor‐intensive rule‐crafting and struggle to adapt automatically to diverse geographic contexts. To alleviate manual effort, hybrid and parametric approaches were introduced, combining exemplar‐based optimization with user‐driven templates to conform generated layouts to real‐world constraints~\cite{aliaga2015inverse}. More recently, data‐driven techniques have gained traction: generative adversarial networks have been applied to synthesize coherent street patterns from satellite imagery~\cite{li2020gan,zhou2022citylayout}, and convolutional neural networks have produced high‐fidelity facade details conditioned on urban morphology~\cite{chen2019facade}. CityDreamer\cite{xie2024citydreamer} employs a compositional generative model to produce unbounded, high‐fidelity 3D city renderings with strong semantic consistency but remains limited to raster outputs without direct vector or CAD integration. However, the prevailing reliance on rasterized or image‐like representations impedes these models from capturing explicit topological relationships and functional dependencies among individual buildings\cite{deng2025citygen}. Emerging graph‐based and relational learning methods aim to bridge this gap by modeling spatial adjacency, land‐use semantics, and connectivity constraints within a unified framework~\cite{gao2021graph,xu2023graphurban}, although these works depends on massive multi-model datasets and hard to tackle with the 3D real world modeling, which are further explored in this paper.

% Early approaches to urban layout generation often relied on procedural modeling techniques, such as shape grammars or rule-based systems~\cite{parish2001procedural}, which offer flexibility but require extensive manual design. More recent methods leverage data-driven models to learn urban patterns from real-world datasets, including GAN-based street layout synthesis~\cite{li2020gan} and CNN-based facade generation. However, these methods typically operate on rasterized representations and struggle to capture relational dependencies between buildings.

% \subsection{Graph-based Modeling in Urban Contexts}
% Graph structures have been increasingly adopted to represent urban environments due to their ability to encode spatial and topological relationships. Works such as OpenStreetGraph \cite{zhou2022openstreetgraph} and UrbanGNN \cite{chen2023urbangnn} demonstrate the effectiveness of graph neural networks in tasks like road network classification and land use prediction. In our work, we extend this paradigm to building-level graph modeling, where nodes represent buildings and edges reflect spatial proximity or functional linkage. This allows for fine-grained control and semantic reasoning in the generative process.

\subsection{Graph-based Modeling in Urban Contexts}

The increasing complexity of urban systems has prompted the adoption of graph‐based representations, where nodes denote spatial entities (e.g., intersections, parcels, buildings) and edges encode topological, semantic, or functional relationships. Foundational graph neural network architectures—such as Graph Convolutional Networks (GCNs) \cite{kipf2017semi} and Graph Attention Networks (GATs) \cite{velivckovic2018graph}—facilitate the propagation of multi‐dimensional features across graph topologies. In the urban domain, these models have been applied to road network classification and embedding via OpenStreetGraph \cite{zhou2022openstreetgraph}, and to land‐use inference with UrbanGNN, which integrates heterogeneous node attributes including point‐of‐interest data and demographic indicators \cite{chen2023urbangnn}. Extending this paradigm to the building level, our work constructs graphs in which nodes represent individual buildings and edges capture spatial adjacency, functional linkage, or pedestrian connectivity. Such granularity enables fine‐grained semantic reasoning and supports tasks ranging from zoning analysis to generative urban design. Moreover, relational extensions of GCNs (R‐GCNs \cite{schlichtkrull2018modeling}) allow the incorporation of multi‐typed edges, accommodating diverse urban relationships such as shared boundaries, infrastructure dependencies, and land‐use synergies. Despite these advances, challenges persist in scaling graph‐based models to city‐wide resolutions and in seamlessly integrating multi‐modal geospatial datasets within unified graph frameworks.

\subsection{Generative Models with Semantic Conditioning}
Generative modeling of structured data has been revolutionized by variational autoencoders (VAEs), which learn latent representations for synthesizing complex objects such as molecular graphs, 3D scenes, and urban layouts. Conditional VAEs (CVAEs) extend this paradigm by injecting auxiliary variables—semantic labels, physical descriptors, or functional constraints—directly into both the encoder and decoder, thereby steering generation toward desired attributes \cite{sohn2015cvae}. To handle arbitrary graph structures, GraphVAE\cite{liu2019cgvae} employs graph neural networks within the VAE framework, embedding node and edge features into a continuous latent space for faithful graph reconstruction and generation \cite{simonovsky2018graphvae}. Complementary hierarchical approaches—such as the Junction Tree VAE (JT-VAE)—decompose molecules into substructure trees, boosting the validity and diversity of generated graphs by modeling both the coarse skeleton and fine-grained connections \cite{jin2018junction}. Inspired by these works, we design a graph-based VAE that incorporates semantic attributes of blocks both the encoder and decoder, enabling semantically consistent and spatially plausible urban building generation.

% Building on these advances, we propose a graph-based CVAE tailored for urban building generation. In our formulation, each node encodes geometry building attributes—footprints, height, size —and each edge captures spatial adjacency or functional linkage. By conditioning the latent distribution on these semantic annotations, our model ensures that the generated urban layouts are both semantically consistent and spatially plausible, facilitating downstream tasks such as zoning analysis, infrastructure planning, and generative urban design.  

% Variational Autoencoders (VAEs) have been widely used for structured data generation, including molecules, scenes, and urban layouts. Conditional VAEs (CVAEs) further enable generation guided by semantic labels or constraints \cite{sohn2015cvae}. For example, CGVAE generates molecular graphs conditioned on chemical properties \cite{liu2019cgvae}. 

\begin{figure*}[t]
\centerline{\includegraphics[width=\textwidth]{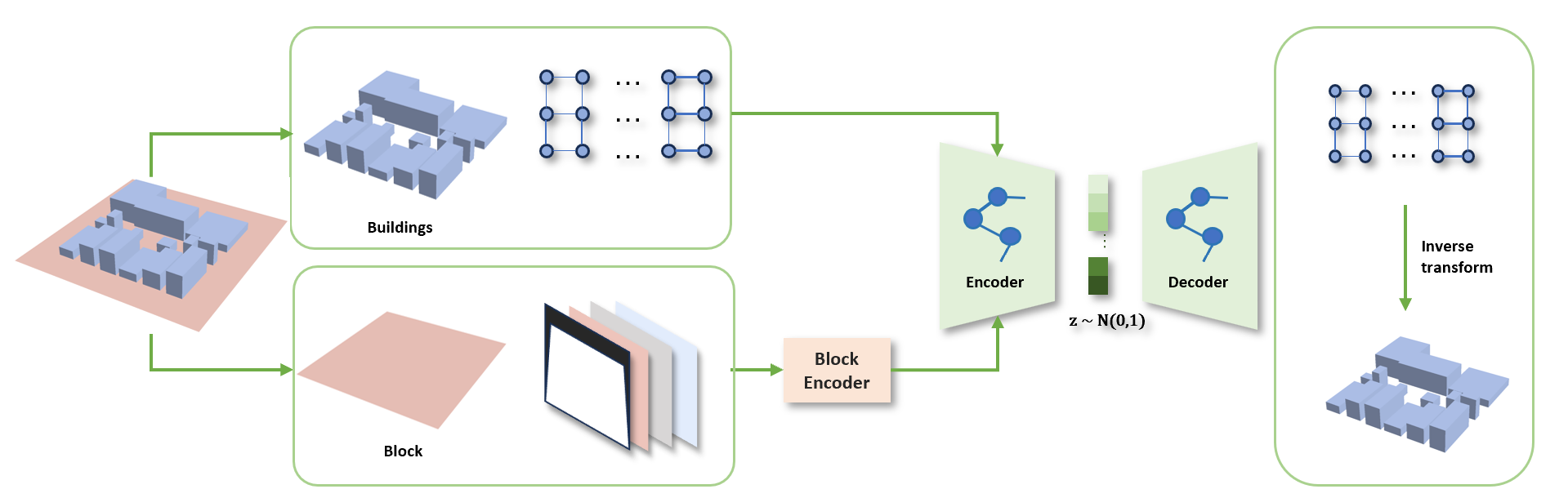}}
\caption{Overall Framework. The CVAE is trained with the canonical graph data of building layout given the block condition containing the semantic and geometric information. }
\label{fig:framework}
\end{figure*}

\section{Our Approach}
The overall workflow of our method is illustrated in the Fig.\ref{fig:framework}. We begin by modeling the spatial arrangement of buildings within a land parcel as a structured graph, which is then fed into a CVAE with a GAT backbone for training. Key attributes of the parcel are concatenated and encoded into a vector that serves as the conditional input to guide the generation process. Once the model is trained, it can generate a 3D urban form by simply providing the boundary of a large-scale parcel along with designated land-use attributes. In the following sections, we first detail the graph representation approach(\ref{subsec:BBR}), followed by an explanation of the CVAE framework(\ref{subsec:CVAE}) and the GAT architecture employed in our model(\ref{subsec:GAT}), finally the block condition encoder will be demonstrated(\ref{subsec:BE}).

\subsection{Block and Building Representation}
\label{subsec:BBR}
We represent an urban block $B$ using two components: \( B = \{b, G\} \), a binary image $b$ that encodes the semantic and geometric attributes of the block, and a graph $G$ that captures the distribution of buildings arranged in multiple rows and columns.

\begin{figure}[h]
    \centering
    \includegraphics[width=0.5\linewidth]{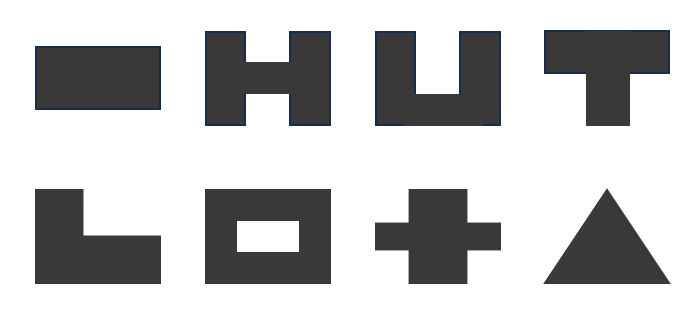}
    \caption{Building shapes library: eight types}
    \label{fig:bldg_shapes}
\end{figure}

\paragraph{Buildings}
A graph can be written as \( G = \{V, E\} \) and each vertice $v_i$ of the set of vertices $V$ describes a single building whose features are defined as: $\mathbf{v}_i = [e_i,\, x_i,\, y_i,\, l_i,\, w_i,\, h_i,\, s_i,\, a_i]$. Building on the geometric normalization transformation proposed by \cite{he2023globalmapper}, where node features describes the existence, normalized positions, normalized size, height, shape feature and iou of specific building respectively. Additionally, we extend the shape fitting library for buildings to include eight distinct parametric forms as shown in Fig.\ref{fig:bldg_shapes}: rectangular, U-shaped, L-shaped, H-shaped, T-shaped, X-shaped, courtyard-type, and triangular.

Additionally, our method introduces a key enhancement by incorporating the adjacency and sparsity characteristics of the building layout. Specifically, edge attributes are defined based on the average inter-row or inter-column distances, enabling variable-weight information propagation between building nodes through the graph structure.For edge $e_{i,j}$ in $E$ ,its edge feature can be calculated by the average distance of adjacent or column:
\begin{align}
e_i &= \frac{\text{dis}(row_{i-1}, row_i)}{W}, \label{eq:row_distance} \\
e_j &= \frac{\text{dis}(col_{j-1}, col_j)}{H}. \label{eq:col_distance}
\end{align}
Here, H and W denote the height and width of the block respectively.

\paragraph{Block}
The block feature $b$ consists of a binary mask $m$, block scale $l$, aspect orientation $p$, and land-use type $u$. While the binary mask $m$ encodes the geometric footprint of the block, the latter three indicators—$l$, $p$, and $u$—are embedded as three additional channels of the mask. Each value of (\( u \in [0,1,2,3] \))  corresponds to a distinct class, and will be transformed via one-hot encoding before being integrated into the model.

These four channels form a multi dimensional representation that captures both the spatial structure and semantic attributes of the urban block, enabling more comprehensive feature encoding during downstream processing.

\subsection{CVAE}
\label{subsec:CVAE}
The Conditional Variational Autoencoder is a generative model that extends the standard VAE by conditioning both the encoder and decoder on auxiliary information \( y \). Given an input \( x \) and condition \( y \), the encoder learns a probabilistic mapping to a latent space \( z \), modeled as \( q_\phi(z|x,y) \), while the decoder reconstructs \( x \) from \( z \) and \( y \) via \( p_\theta(x|z,y) \).

The CVAE is trained by maximizing the conditional evidence lower bound (ELBO), which balances the reconstruction accuracy and the regularization of the latent space. The objective function is defined as:
\begin{align}
\mathcal{L}_{\text{CVAE}}(\theta, \phi; x, y) 
&= \mathbb{E}_{q_\phi(z|x,y)}[\log p_\theta(x|z,y)] \notag \\
&\quad - \mathrm{KL}(q_\phi(z|x,y) \| p(z|y)) \label{eq:cvae_loss}
\end{align}

Here, the first term encourages accurate reconstruction of the input conditioned on \( y \), and the second term is the Kullback–Leibler divergence that regularizes the approximate posterior \( q_\phi(z|x,y) \) to be close to the prior \( p(z|y) \), typically chosen as a standard Gaussian independent of \( y \).

By incorporating the condition \( y \) into both encoding and decoding processes, CVAE enables controlled generation and structured representation learning, making it well-suited for tasks involving attribute-guided synthesis or disentangled feature modeling.

\subsection{Graph Attention Network}
\label{subsec:GAT}
%-------------------------------------------------------------------
% A concise description of GAT with edge-weighted attention
%-------------------------------------------------------------------

% \subsection*{Graph Attention Networks with Edge Weights}
GAT enable flexible node representation learning by aggregating information from neighboring nodes using attention mechanisms. For each node \(v_i\), its feature \(\mathbf{h}_i\) is first linearly transformed:
\begin{align}
\mathbf{h}_i' &= \mathbf{W}\,\mathbf{h}_i
\end{align}
An attention score measures the influence of neighbor \(v_j\):
\begin{align}
e_{ij} &= \mathrm{LeakyReLU}\left(\mathbf{a}^\top[\mathbf{h}_i' \Vert \mathbf{h}_j']\right)
\end{align}
To incorporate edge-specific relationships—such as spatial proximity or semantic similarity—a weight \(w_{ij}\) modulates the score:
\begin{align}
    \tilde{e}_{ij} &= w_{ij} \cdot e_{ij}
\end{align}
These modulated scores are normalized via softmax:
\begin{align}
    \alpha_{ij} &= \frac{\exp(\tilde{e}_{ij})}
               {\sum_{k \in \mathcal{N}_i}\exp(\tilde{e}_{ik})}
\end{align}
Finally, the updated feature for node \(v_i\) is a weighted sum of its neighbors:
\begin{align}
    \mathbf{h}_i'' &= \sigma\left(\sum_{j \in \mathcal{N}_i}\alpha_{ij}\,\mathbf{h}_j'\right) 
\end{align}

By integrating edge weights into the attention mechanism, GAT adaptively emphasizes more informative or spatially relevant neighbors during feature propagation.

\subsection{Block Encoder}
\label{subsec:BE}
To extract meaningful representations of urban blocks, we design a block encoder that processes the four-channel feature image. A convolutional neural network-based encoder is employed to process the input multi-channel image, which learns to compress the spatial and semantic information into a compact latent vector. The encoder is trained by minimizing the reconstruction loss. The resulting feature embedding is a 128-dimensional vector set in our experiment, which proves to be a concise and expressive descriptor of the block's geometry and semantics.

\section{Experiments}

\subsection{Dataset}
To construct a comprehensive dataset for urban block modeling, we collected and integrated data from multiple sources. Building footprints with height information were obtained from OpenStreetMap (OSM) and the Microsoft Building Footprints dataset \cite{osm,msbf}. Block-level boundaries were downloaded from the TIGER/Line Shapefiles provided by the U.S. Census Bureau \cite{tiger}. Additionally, land-use type data for cities including New York, Chicago, and Washington, D.C. were retrieved from their respective official urban data portals \cite{nycdata,chicagodata,dcdata}.

All raw data sources were provided in vector format (e.g., shapefiles or GeoJSON), which facilitated spatial alignment and integration. After preprocessing and merging, the final dataset comprises approximately 5,000 urban blocks and 25,000 individual building footprints, each annotated with geometric and semantic attributes. A summary of the dataset statistics is presented in Table~\ref{tab:bldg_num} and Table~\ref{tab:landuse}, including counts, coverage, and attribute completeness across different cities.

\begin{table}[h]
\caption{Summary of building counts per block in each city}
\label{tab:bldg_num}
\centering
\begin{tabular}{lccccc}
\toprule
City & Max & Min & Avg & std & Total Blocks \\
\midrule
NYC & 392 & 1 & 45.6 & 21.2 & 62535\\
Chgo & 460 & 1 & 17.7 & 15.6 & 46357 \\
LA & 286 & 1 & 28.2 & 25.6 & 23418 \\
\bottomrule
\end{tabular}
\end{table}

\begin{table}[h]
\caption{Number of blocks per land use type in each city}
\label{tab:landuse}
\centering
\begin{tabular}{lccc}
\toprule
Landuse & NYC & Chgo & LA \\
\midrule
Residential & 14674 & 4991 & 4857 \\
Commercial & 6233 & 6465 & 4325 \\
Industrial & 1425 & 851 & 699 \\
Public & 3028 & 285 & 396 \\
Recreation & 7538 & 599 & 481 \\
\bottomrule
\end{tabular}
\end{table}

\begin{figure*}[t]
    \centering

    % 第一张子图
    \begin{subfigure}[t]{0.24\textwidth}
        \includegraphics[width=\linewidth]{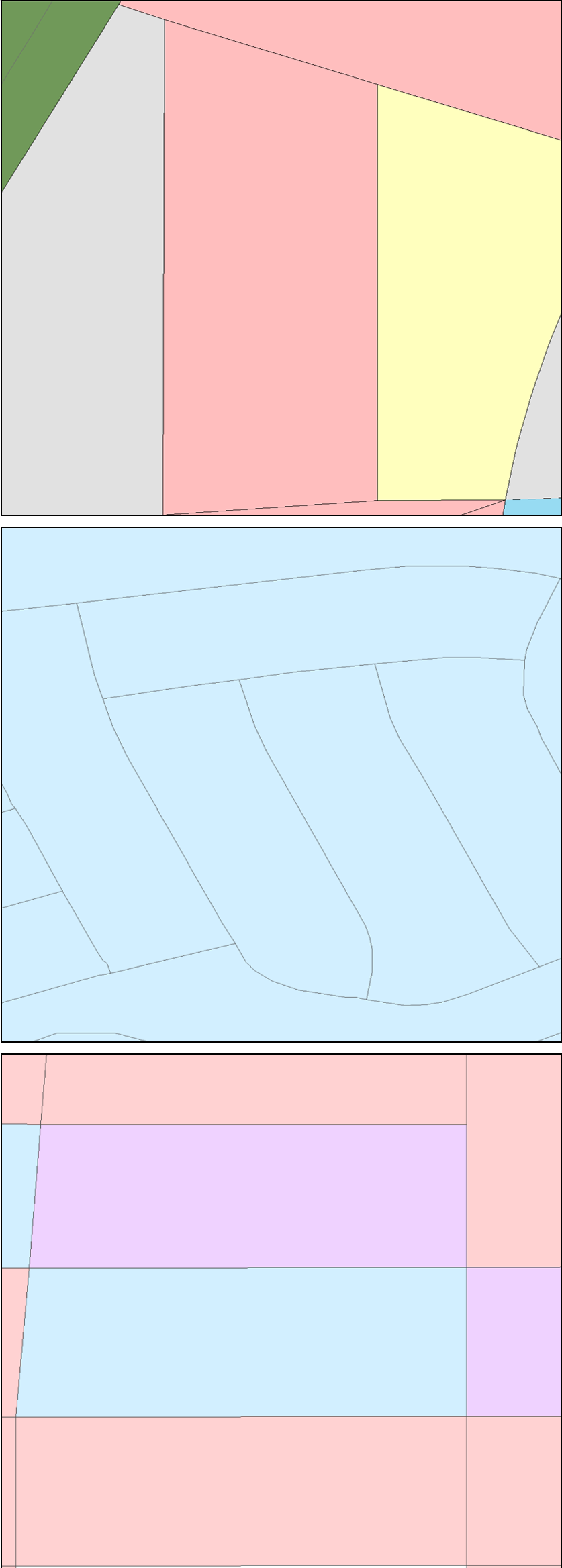}
        \caption{InputBlock}
    \end{subfigure}
    \hfill
    % 第二张子图
    \begin{subfigure}[t]{0.24\textwidth}
        \includegraphics[width=\linewidth]{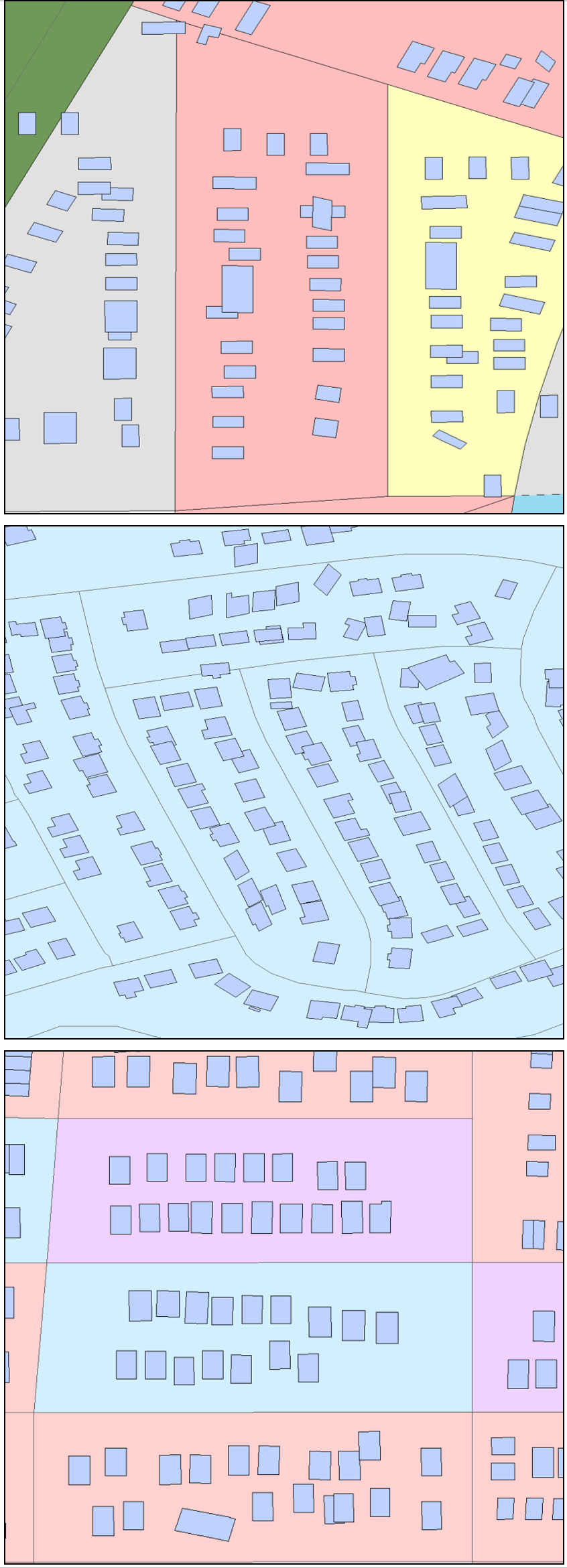}
        \caption{BlockPlanner}
    \end{subfigure}
    \hfill
    % 第三张子图
    \begin{subfigure}[t]{0.24\textwidth}
        \includegraphics[width=\linewidth]{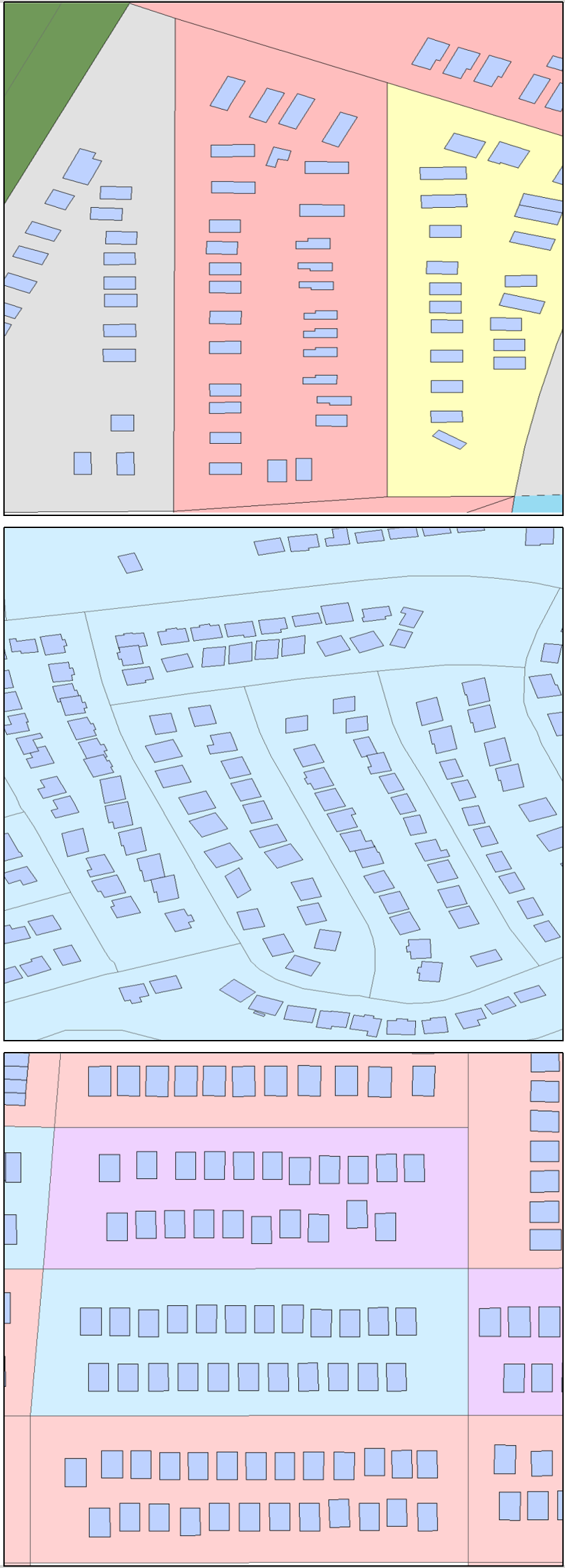}
        \caption{GlobalMapper}
    \end{subfigure}
    \hfill
    % 第四张子图
    \begin{subfigure}[t]{0.24\textwidth}
        \includegraphics[width=\linewidth]{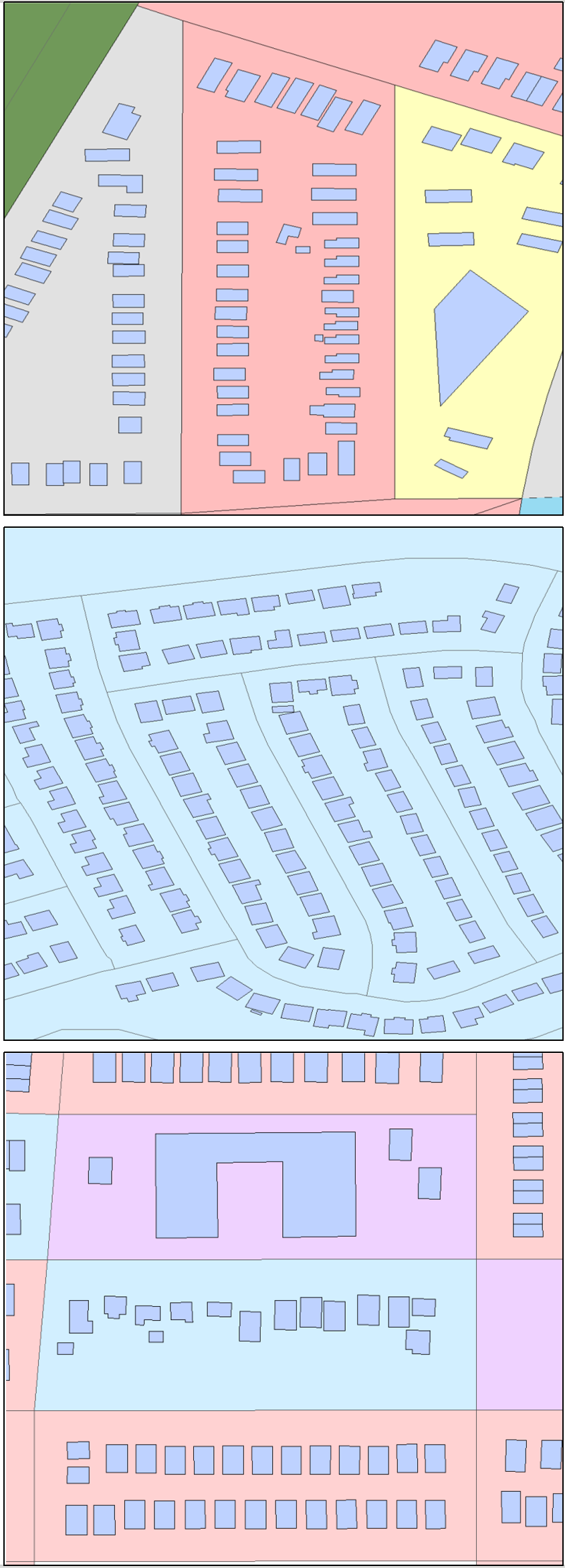}
        \caption{Ours(2d)}
    \end{subfigure}

    \caption{Given parcel boundaries and land-use types(represented by different colors), the generated 2D building layouts are shown. BlockPlanner\cite{xu2021blockplanner} often produces overlapping buildings and violates parcel constraints. GlobalMapper\cite{he2023globalmapper} lacks semantic awareness, resulting in repetitive and homogeneous layouts. Compared to these methods, our approach generates diverse, semantically consistent building configurations with reduced spatial error.}
    \label{fig:result}
\end{figure*}

% \begin{figure*}[t]
%     \centering
%     \includegraphics[width=\linewidth]{figures/2d_total.png}
%     \caption{City Layout within different regions.The first row illustrates 3D building layouts in predominantly commercial areas; the second row presents generation results for mixed commercial, residential, and industrial zones as well as irregularly shaped plots; and the third row shows outcomes for regions with regular plots and mixed land‐use types.}
%     \label{fig:result}
% \end{figure*}

\subsection{Implementation Details}
We conducted end-to-end training following the proposed methodology, using approximately 20,000 urban parcels sampled from three major cities, with each city contributing roughly one-third of the total training data. The model typically converged after around 200 iterations. For supervision, we applied a cross-entropy loss to the newly introduced node height attribute, with a weighting factor of 4. Additionally, the edge weights representing inter-row and inter-column spacing were trained using a cross-entropy loss with a weight of 1. The total number of training epochs was set to 250. All experiments were performed on a single NVIDIA RTX 4090 Ti GPU, with the complete training process taking approximately 48 hours.

\subsection{Main Results}
Given a target urban area and specified land-use types for each block, our model can automatically generate corresponding 3D building layouts within the designated parcels. As illustrated in the ~\ref{fig:result}, we showcase mixed-use blocks with varying land-use configurations to demonstrate the model's ability to synthesize building arrangements conditioned on both geometric and semantic attributes of the input parcels. The first row shows 3D layouts in primarily commercial districts, marked by symmetrical, axis-aligned plots and largely uniform building heights. The second row presents results for mixed-use areas—commercial, residential, and industrial—and for irregularly shaped parcels: industrial buildings are sparsely distributed, residential blocks follow consistent patterns, and commercial structures are larger, taller, and more irregular to suit their functions. The third row illustrates regions with regular plot geometries and mixed land uses, where buildings are arranged in a clear row-and-column grid.

Notably, even within the same geographic region, altering the land-use type alone enables the generation of diverse urban layout schemes, as shown in Fig \ref{fig:urban_layout}. The outputs are structured as vector geometries, making them readily editable and suitable for downstream optimization and planning workflows.

\begin{table*}[htbp]
\centering
\caption{We generate 500 urban layouts and compare to the same amount of real urban layouts. Best values are in \textbf{bold}. Our method outperforms other existing methods in 3 metrics.}
\label{tab:quantitative_results}
\begin{tabular}{lcccccc}
\hline
\textbf{Method} & \textbf{L-Sim↑} & \textbf{OPR↓ (\%)} & \textbf{OBR↓ (\%)} & \textbf{FID↓} & \textbf{WD↓ (bbx)} & \textbf{WD↓ (count)} \\
\hline
BlockPlanner & 10.84  & 4.25  & 1.26 & 26.77 & 6.23      & \textbf{0.15} \\
GlobalMapper & 12.36 & \textbf{0.98}         & 0.45 & 15.32 & \textbf{2.52} & 0.38        \\
Ours              & \textbf{12.96} & 1.03 & \textbf{0.37} & \textbf{11.32} & 2.7 & 0.4 \\
\hline
\end{tabular}
\end{table*}

\subsection{Evaluation Metrics}
To quantitatively evaluate the quality of generated building layouts, we adopt a set of spatial metrics that capture geometric accuracy, distributional similarity, and topological validity. Specifically, we compute:

\begin{itemize}
    \item \textbf{Layout Similarity (L-Sim)}: Measures the spatial similarity between generated and reference layouts using Hausdorff or Earth Mover's Distance.
    \item \textbf{Overlap Ratio(OPR)}: Quantifies the proportion of overlapping area among buildings.
    \item \textbf{Out-of-Block Ratio(OBR)}: Evaluates the percentage of buildings that fall outside their designated block boundaries.
    \item \textbf{Wasserstein Distance (Bounding Boxes)}: Assesses the distributional difference of building positions and sizes via bounding box centers.
    \item \textbf{Wasserstein Distance (Counts)}: Compares the per-block building count distributions between generated and reference layouts.
    \item \textbf{Fréchet Inception Distance (FID)}: Optionally used to measure visual similarity by rasterizing layouts and comparing feature distributions extracted from a pretrained network.
    % \item \textbf{Diversity (DIV)}: the mean intersection-over-union(mIoU) between the generated results and the real results.
\end{itemize}

These metrics provide a comprehensive assessment of both structural fidelity and semantic plausibility in urban layout generation.

\subsection{Comparison}
Based on the evaluation metrics described above, we compared our building layout generation method with existing graph-based approaches, including GlobalMapper\cite{he2023globalmapper} and BlockPlanner\cite{xu2021blockplanner}. A total of 500 urban blocks were selected for quantitative analysis. Since these baseline methods do not incorporate building height information, the comparison was conducted primarily on two-dimensional spatial distributions. As shown in Table~\ref{tab:quantitative_results}, our method outperforms the other two approaches on three key metrics, and achieves comparable performance on Layout Similarity and Out-of-Block Ratio relative to the best results.

\begin{figure*}[htbp]
    \centering
    \includegraphics[width=0.8\linewidth]{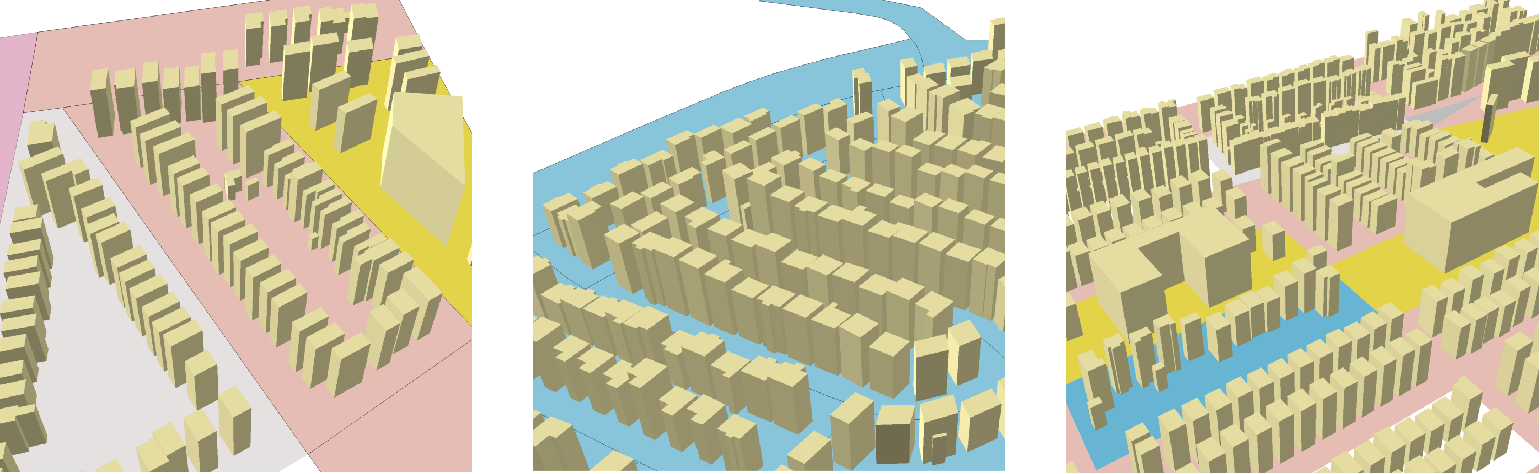}
    \caption{The 3D layout generation samples.}
    \label{fig:3Dres}
\end{figure*}

Additionally, our method support the 3D fundamental buildings layout generation as shown in \ref{fig:3Dres}, which can be easily utilised in downstream applications due to its vector-based 3D data structure.

From a broader perspective, we observe that extending the building shape library to include more complex geometric primitives does not significantly improve the similarity between generated layouts and ground truth distributions. In contrast, the incorporation of land-use types and edge weights appears to enhance the structural coherence of grid-based building arrangements, making it easier to uncover latent spatial patterns informed by semantic attributes. While all compared methods support the generation of planar vector data, our approach additionally enables the generation of vector-based layouts with 3D height attributes and supports editable land-use configurations.

\section{Conclusion}
We propose a controllable method for generating large-scale 3D vector-based urban layouts. By leveraging road networks or block boundaries along with semantic information, our approach enables the creation of diverse urban design schemes with functional attributes. Compared to existing methods, our results demonstrate effective performance and enhanced flexibility.

\subsection{Application and Limitation}
This method offers valuable support for professionals in urban planning and related fields. The generated layouts are editable and can serve as foundational models for more detailed urban design tasks.

However, the approach has limitations. Representing buildings as simple 3D extrusions (cylindrical forms) fails to capture the complexity of architectural structures and surrounding environments. While the integration of semantic information facilitates designer involvement, it also introduces challenges in attribute specification and consistency.

\subsection{Further Work}
Future research will explore hybrid approaches that combine automated generation with human-in-the-loop refinement. This includes producing diverse urban layouts with functional zoning, automated evaluation and optimization feedback, and allowing designers to intervene during the process. We also aim to enhance the granularity of 3D models and incorporate additional urban elements such as vegetation and water bodies to better simulate organic urban environments. These directions align with emerging efforts in large language models and computer vision applications.

\begin{acks}
Thanks to everyone who has provided their valuable suggestions and all the members in our group.
\end{acks}

\bibliographystyle{ACM-Reference-Format}
\bibliography{sample-base}

\appendix









\end{document}